\documentclass[runningheads]{llncs}

\usepackage[year=2026]{eccv}
\usepackage{eccvabbrv}

\usepackage{graphicx,amsmath,amssymb,booktabs,xcolor,colortbl}
\usepackage{placeins}
\usepackage{tikz}
\usetikzlibrary{arrows.meta,positioning,fit}

\usepackage{hyperref}
\hypersetup{hidelinks}
\usepackage[capitalize]{cleveref}
\begin{document}

\title{Rate-Adaptive One-Step Diffusion Compression for AIGC Images}
\titlerunning{Rate-Adaptive Diffusion Compression for AIGC Images}

\author{Nitiz Khanal}
\authorrunning{N. Khanal}

\institute{Pulchowk Campus, Institute of Engineering, Tribhuvan University, Lalitpur, Nepal \\
\email{077bel025.nitiz@pcampus.edu.np}}

\maketitle

\begin{abstract}
We describe our entry to the LoViF 2026 AIGC Image Compression Challenge, a benchmark for ultra-low-bitrate coding of AI-generated images under a strict global rate budget of 0.025 bits per pixel (BPP). Generated imagery poses a distinct challenge for compression: it frequently contains rendered typography, synthetic edges, repeated motifs, UI-like layout, and stylized micro-texture that conventional distortion-oriented codecs erase at this rate, while unconstrained generative decoders can restore plausible-looking detail that no longer matches the source geometry or symbols. We treat this as a rate-perception allocation problem. Our system fine-tunes four rate-specialized checkpoints of the AEIC one-step diffusion codec, generates per-image candidates from all four, including one latent refined through encoder-side test-time optimization (TTO) with periodic entropy-conditioning refresh, entropy-codes every candidate with practical rANS coding, and selects exactly one bitstream per image with a multiple-choice-knapsack MILP over measured coded sizes. A fixed, zero-additional-bit residual restoration network is applied at decode time. Every submitted bitstream is independently decodable by the shipped decoder, which uses no source image or external side information. We report the full pipeline, an ablation history spanning 77 logged development experiments, and a set of negative results, including a systematic investigation of why PSNR could not be pushed to parity with rate-distortion-oriented competitors under this architecture, that we believe are useful to future participants. \emph{This is a challenge paper: it reports our entry to the LoViF 2026 AIGC Image Compression Challenge, which scored 31.527739 (PSNR 27.02\,dB, MS-SSIM 0.9176, LPIPS 0.0778, DISTS 0.0390 at 0.02495 BPP), achieving the second-best DISTS on the leaderboard and ranking 5th on the organizer-recalculated final test-phase leaderboard announced August 4, 2026.}
\end{abstract}

\section{Introduction}
The LoViF 2026 AIGC Image Compression Challenge studies ultra-low-bitrate coding under a strict global rate constraint (average BPP $\leq 0.025$), with distortion (PSNR), structural similarity (MS-SSIM), and two learned perceptual metrics (LPIPS, DISTS) evaluated jointly. This setting is unusually demanding for AI-generated content: generated images routinely contain rendered typography, synthetic edges, repeated motifs, UI-like layouts, and stylized texture that a conventional rate-distortion-optimized codec removes almost entirely below 0.03 BPP, while an unconstrained generative decoder can hallucinate plausible replacement detail that changes symbol identity or edge geometry and is therefore penalized by every one of the four metrics simultaneously.

Our approach frames this as a rate-perception allocation problem rather than a single-operating-point compression problem. An entropy-coded latent from a fine-tuned learned codec preserves compact structural evidence about the source image; a frozen one-step diffusion prior (SD-Turbo~\cite{sauer2023sdturbo}) supplies a strong natural-image manifold that the latent conditions; and encoder-side optimization adapts the latent to each source image without ever modifying decoder weights. Because a single rate/perception operating point is not uniformly best across a heterogeneous AIGC test set, we further maintain four rate-specialized checkpoints and select, per image, whichever of several real coded candidates maximizes the challenge score with a multiple-choice-knapsack MILP over measured (not estimated) bitstream sizes. The rate accounting is exact: the constraint uses actual file sizes and total pixels. This is explicitly a \emph{dataset-level, offline} allocation setting: the encoder has the complete fixed image set and its sources before selection. It is appropriate to the challenge's aggregate-BPP rule, but is not an online controller and does not provide independent per-image rate guarantees.

\paragraph{Contributions.} (1) A practical rate-adaptive pipeline built on a one-step diffusion codec (AEIC-ME) with four rate-specialized checkpoints, encoder-side latent TTO with periodic entropy-conditioning refresh, and a fixed zero-additional-bit decoder-side residual restorer. (2) A multiple-choice-knapsack MILP over \emph{real, file-size-measured} bitstreams (not per-image mean BPP or entropy estimates) for global rate allocation; fixed per-file overhead and heterogeneous image resolutions require a total-bit constraint rather than a mean of per-image BPP values. (3) A documented ablation history of 77 development experiments, including several informative negative results: cross-family codec fusion with a VQ-latent codec (OneDC) and a second one-step diffusion codec (OSCAR) were both rejected as non-competitive; a directly differentiable PSNR TTO objective did not close the PSNR gap to rate-distortion-oriented competitors, suggesting the gap may reflect the architecture's fidelity/perception trade-off rather than the loss weighting used to steer it.

\section{Related Work}
\label{sec:related}
\paragraph{Learned image compression.} Modern learned codecs pair a nonlinear analysis/synthesis transform with a learned entropy model, typically a scale or mean-scale hyperprior~\cite{balle2018hyperprior} combined with an autoregressive or channel-wise context model~\cite{he2022elic}, and use a practical range coder such as an asymmetric numeral system (rANS)~\cite{duda2009rans} to realize the bitstream. These systems are distortion-oriented: at ultra-low rate they minimize pixel or feature error subject to a rate penalty, which tends to blur or remove fine, high-frequency structure such as text and repeated patterns rather than hallucinate it.

\paragraph{Diffusion priors for extreme compression.} Recent work augments the synthesis transform with a generative prior to recover perceptually plausible detail at rates where distortion-oriented decoders visibly blur. One-step distilled diffusion models such as SD-Turbo~\cite{sauer2023sdturbo}, distilled from latent diffusion backbones~\cite{rombach2022ldm}, make a diffusion-conditioned synthesis path practical at single-feed-forward inference cost. Our base codec, AEIC~\cite{zhang2026aeic}, extends its predecessor StableCodec~\cite{zhang2025stablecodec} (which reaches 0.005 BPP) with a shallower encoder and rate-specialized fine-tuning. Several concurrent challenge-era systems target comparable sub-0.03 BPP operating points via the same one-step-diffusion idea: OneDC~\cite{xue2025onedc}, OSCAR~\cite{guo2025oscar}, GLC~\cite{qi2025glc}, DiffO~\cite{park2025diffo}, SPRDiff~\cite{wei2026sprdiff}, ResULIC~\cite{ke2025resulic}, and a dual-latent decoder~\cite{mao2026duallatent}. We evaluated OSCAR and OneDC directly (\cref{sec:ablation}), on the release available 2026-07-13/14. Screening OSCAR's official checkpoint on 12 validation images across its three published rate tiers gave board-scale scores of $-11.56$, $2.44$, and $16.14$ (at 0.0019/0.0098/0.0313 BPP), all far below our 31.53; its \texttt{z\_only} inference path also reports a fixed-length VQ-index bit count rather than invoking its own compiled rANS coder, so no real serialized bitstream results at that operating point. OneDC's \texttt{exlow\_bpp0034} checkpoint showed the same limitation in the release we evaluated, and, on a 6-image screen limited to sub-1K-pixel images by an OOM in its untiled VAE decode, scored far below AEIC (PSNR 16--20\,dB, LPIPS 0.24--0.40, DISTS 0.13--0.23) at a much lower, unmatched native rate (0.0034 BPP vs.\ our 0.025 BPP submission).

\paragraph{AIGC-specific and text-aware compression.} Because AI-generated images disproportionately contain rendered text, UI chrome, and repeated stylized motifs relative to natural-image benchmarks, concurrent efforts such as TextBoost~\cite{wang2026textboost} bias compression toward text legibility using OCR side-information or attention-guided fusion at ultra-low rate. Our system transmits no OCR side-information (EasyOCR is used only at the encoder, \cref{sec:tto}); instead, the base codec (AEIC) is itself trained with an overlap-chunked, edge-aware DISTS objective intended to preserve exactly this class of structure, and our fine-tuning and TTO stages inherit that inductive bias rather than adding a separate text-specific branch.

\paragraph{Perceptual quality metrics.} We optimize against and report the same four metrics used by the challenge: PSNR, multi-scale structural similarity (MS-SSIM)~\cite{wang2003msssim}, LPIPS~\cite{zhang2018lpips} (AlexNet backbone, matching the organizers' evaluator), and DISTS~\cite{ding2020dists}, the latter two being learned metrics computed from deep features rather than pixelwise error; PSNR and MS-SSIM are computed in RGB, and our local evaluator matches the organizers' displayed leaderboard values digit-for-digit on every submission checked (\cref{sec:eval}). Our supporting learned-compression infrastructure (rate-distortion training utilities, entropy modules) draws on CompressAI~\cite{begaint2020compressai}.

\section{Method}
\subsection{Overview}
\begin{figure}[t]
\centering
\resizebox{\textwidth}{!}{%
\begin{tikzpicture}[font=\small, node distance=7mm and 9mm,
 box/.style={draw,rounded corners,align=center,minimum height=9mm,minimum width=23mm,fill=blue!5},
 arr/.style={-{Latex[length=2mm]},thick}]
\node[box] (src) {source\\image $x$};
\node[box,right=of src] (multi) {four rate-specialized\\analysis transforms};
\node[box,right=of multi] (tto) {per-image latent TTO\\frozen decoder/entropy context};
\node[box,right=of tto] (rans) {quantization +\\rANS coding};
\node[box,right=of rans] (pool) {candidate streams\\and measured rates};
\node[box,below=12mm of pool] (knap) {global MILP rate allocation\\$\sum_i |b_i|/\sum_i H_iW_i\leq0.025$};
\node[box,left=of knap] (stream) {selected bitstream\\+ checkpoint ID};
\node[box,left=of stream] (dec) {entropy decode +\\one-step diffusion};
\node[box,left=of dec] (enh) {fixed residual\\restoration CNN};
\node[box,left=of enh] (out) {reconstruction\\$\hat{x}$};
\draw[arr] (src)--(multi); \draw[arr] (multi)--(tto); \draw[arr] (tto)--(rans);
\draw[arr] (rans)--(pool); \draw[arr] (pool)--(knap); \draw[arr] (knap)--(stream);
\draw[arr] (stream)--(dec); \draw[arr] (dec)--(enh); \draw[arr] (enh)--(out);
\draw[arr,dashed] (src.south) |- (knap.west);
\end{tikzpicture}}
\caption{Final compression pipeline. Dashed flow denotes encoder-only access to the source for candidate evaluation and rate allocation; the submitted decoder receives only the selected bitstream and reconstructs without any source access.}
\label{fig:pipeline}
\end{figure}
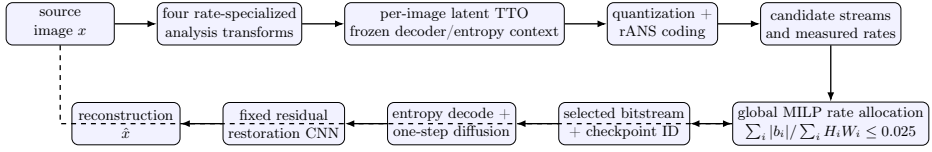

\Cref{fig:pipeline} summarizes the submitted system. Four rate-specialized AEIC-ME checkpoints share one decoding topology but occupy different rate/perception operating points. For each test image, the encoder evaluates admissible raw candidates from all four checkpoints plus one TTO-refined candidate, entropy-codes each with practical rANS coding, and records a one-byte checkpoint identifier alongside the coded payload. The candidate pool also contains raw and restored variants, but that restoration choice is not encoded in the bitstream. A global multiple-choice-knapsack MILP then selects exactly one candidate stream per image, subject to the true aggregate bit budget computed from real coded file sizes and true image dimensions. At decode time the checkpoint identifier selects the matching codec, a tiled one-step diffusion pass reconstructs the image from the decoded latent, and a fixed, zero-additional-bit residual restoration network produces the final RGB output. The decoder performs no test-time optimization and requires no access to the source image.

\subsection{Backbone: AEIC-ME with a one-step diffusion prior}
\label{sec:backbone}
The base codec is the ME variant of AEIC~\cite{zhang2026aeic}, a one-step diffusion image codec whose analysis side maps RGB content into a compact quantized latent governed by a learned autoregressive entropy model, and whose synthesis side conditions an SD-Turbo one-step diffusion network~\cite{sauer2023sdturbo} on the decoded latent. SD-Turbo supplies a high-capacity natural-image prior while avoiding the sampling-step cost of the multi-step latent diffusion models it is distilled from~\cite{rombach2022ldm}. Our decoder bundle uses the trimmed FP16 SD-Turbo UNet and VAE together with the AdcSR half-decoder used by the AEIC implementation for the final pixel-space decode; large images are reconstructed using overlapping latent and VAE tiles, with overlap exceeding the effective local receptive boundary so that seams are not visible while peak activation memory stays bounded on a single consumer/cloud GPU.

\subsection{Rate-specialized checkpoint bank and bitstream signaling}
Zero-shot AEIC checkpoints are tuned for a broad rate ladder rather than for this challenge's exact 0.025 BPP AIGC operating point (\cref{tab:e1e3}). We fine-tuned four checkpoints from the public AEIC-ME lineage, varying the loss weights, initialization, and training corpus to cover complementary rate/perception regions. Fine-tuning freezes the VAE/AdcSR decoder and the base UNet weights, but updates the complete AEIC codec together with UNet LoRA parameters and its input/output convolutions. Every checkpoint shares the same decoder entry point: the encoder writes one leading byte selecting the checkpoint, and the remainder is decoded by the shared practical entropy coder. No additional flag records whether the selected candidate was raw or passed through the restorer, so the standalone decoder applies the restorer unconditionally. Checkpoint selection is therefore reproducible from the bitstream alone, while the raw/restored candidate choice is not. \Cref{tab:ckpts} documents the four checkpoints' training configuration. \Cref{tab:ckptmetrics} reports their retained full-test metrics, restoration deltas, and final bitstream selection counts; all four specialists contribute to the submitted allocation.

\begin{table}[t]
\centering\scriptsize
\begin{tabular}{@{}lllcc@{}}
\toprule
Checkpoint & Init & Steps & Corpus & Target rate \\
\midrule
aigc8\_5000 & ft8 (zero-shot, $\lambda_8$) & 5k & train-800 $8\times$ + initial synth. & 0.020 BPP \\
r2\_18000 & aigc4\_5000 ($\lambda_4$) & 16k+ & train-800 $8\times$ + synth. & 0.037 BPP \\
r3l4\_8000 & r2\_18000 & 8k & train-800 $8\times$ + SD3.5/SDXL synth. & 0.037 BPP \\
r3l8\_8000 & aigc8\_5000 & 8k & train-800 $8\times$ + SD3.5 synth. & 0.026 BPP \\
\bottomrule
\end{tabular}
\caption{Training configuration of the four shipped checkpoints. All use AEIC's codec loss $\lambda_RR+\lambda_2\|x{-}\hat x\|_2^2+\lambda_D\mathcal{L}_{\mathrm{DISTS}}+\lambda_C\mathcal{L}_{\mathrm{CLIP}}+\lambda_A\mathcal{L}_{\mathrm{adv}}+\lambda_L\mathcal{L}_{\mathrm{LPIPS}}$. Exact weights $(\lambda_R,\lambda_2,\lambda_D,\lambda_C,\lambda_A,\lambda_L)$ are $(8,2,1,.1,.1,1)$ for aigc8\_5000; $(4,2,1.5,.1,.1,2)$ for r2\_18000 and r3l4\_8000; and $(8,2,1.5,.1,.1,2)$ for r3l8\_8000. These values are retained in the four released training configuration files. Target rate is the checkpoint's measured natural operating region, not an imposed per-image constraint.}
\label{tab:ckpts}
\end{table}

\begin{table}[t]
\centering\tiny
\resizebox{\textwidth}{!}{%
\begin{tabular}{@{}lrrrrrrrrrr@{}}
\toprule
& \multicolumn{5}{c}{Raw checkpoint} & \multicolumn{4}{c}{Restorer change} & Final \\
\cmidrule(lr){2-6}\cmidrule(lr){7-10}
Checkpoint & BPP & PSNR & MS-SSIM & LPIPS & DISTS & $\Delta$PSNR & $\Delta$MS-SSIM & $\Delta$LPIPS & $\Delta$DISTS & selected \\
\midrule
aigc8\_5000 & .019739 & 25.6427 & .891992 & .112686 & .063923 & +.1848 & +.000272 & -.001876 & -.001634 & 20 \\
r2\_18000   & .036242 & 27.0023 & .921605 & .082632 & .049385 & +.2166 & +.000084 & -.001581 & -.001376 & 12 \\
r3l4\_8000  & .036478 & 27.0471 & .922338 & .081227 & .048786 & +.2114 & +.000063 & -.001420 & -.001200 & 40 \\
r3l8\_8000  & .025387 & 26.3583 & .907997 & .091023 & .052620 & +.1948 & +.000164 & -.001377 & -.001219 & 28 \\
\bottomrule
\end{tabular}}
\caption{Per-checkpoint full-test results before global selection, retained from the actual coded candidates. These are descriptive measurements of frozen, already-generated challenge candidates and were not used to update checkpoint weights. Restorer columns are restored-minus-raw metric changes (negative is better for LPIPS/DISTS); BPP is unchanged. ``Final selected'' counts the checkpoint-ID bytes in the 100 submitted streams, including streams whose latent was refined by TTO.}
\label{tab:ckptmetrics}
\end{table}

\subsection{Encoder-side latent test-time optimization}
\label{sec:tto}
For the rf500 candidate (the only TTO-derived member of the 10-tag final knapsack pool, \cref{sec:knapsack}), TTO refines the continuous pre-quantization latent $y$ with Adam (lr $10^{-3}$, no schedule) for 500 iterations (earlier development used 300; \cref{sec:ablation} reports that history), freezing all decoder weights and, within each refreeze interval (below), the entropy model's autoregressive conditioning. Quantization uses a straight-through estimator, $\hat y=\mathrm{STE\text{-}round}(y-\mu)+\mu$. Each iteration samples one random latent crop ($LY{=}16$, i.e.\ 512 source px), decodes it through the frozen one-step diffusion path (no restorer in this loop; see below), and computes
\begin{equation}
 \mathcal{L}_{\mathrm{TTO}}=\lambda_2\|x-\hat{x}\|_2^2+40\,\mathcal{L}_{\mathrm{LPIPS}}+\lambda_D\mathcal{L}_{\mathrm{DISTS}}+\lambda_R R,
 \label{eq:tto}
\end{equation}
with $\lambda_2{=}1000$ (\texttt{mse\_w}), $\lambda_D{=}40$ (\texttt{dists\_w}), $\lambda_R{=}20000$ (\texttt{rate\_w}), and the LPIPS coefficient fixed at 40. $R=\mathrm{ReLU}(\hat{r}-r_0)$ is a hinge penalty against the trajectory's own \emph{starting} rate $r_0$ (stop-gradient), so TTO is only penalized for exceeding where it began, not driven toward a target rate; $\hat r$ is the entropy model's differentiable rate \emph{estimate} (negative log-likelihood of $y$ under the frozen Gaussian conditional, in bits per pixel); this is a standard soft proxy, \emph{not} the true rANS bit count, since arithmetic coding is not differentiable. This estimate is used only to steer the trajectory; after every trajectory the final $y$ is re-quantized and passed through the actual, non-differentiable rANS entropy encoder, so every reported/scored/legal-budget bit count in this paper (\cref{tab:ablation,tab:results,tab:leaderboard}) is a measured file size, never the estimate. One finished, fixed-iteration trajectory per image is handed to the final knapsack, with no random restarts or early stopping.

\paragraph{Text-guidance ablation.} OCR guidance is \emph{not} used by the rf500 trajectory in the final test archive. We reran a matched 100-step ablation on the 52/100 validation images where EasyOCR found text, using the same four-checkpoint seed plan, optimization settings, and random seed. The guided arm sampled a detected-text crop with probability 0.7 and weighted DISTS $3\times$ on those iterations; the control used ordinary random crops. Whole-image performance did not improve: plain TTO scored $S{=}110.0559$ at 0.032243 BPP, versus $110.0050$ at 0.032257 BPP with guidance ($-0.0509$ $S$; 24/52 images improved). Direct crop analysis covered 392 source crops with nonempty source OCR text across 51 images. Guidance changed text-region PSNR by $-0.068$\,dB, LPIPS by $-0.00011$, and DISTS by $-0.00154$ (lower is better). Using normalized EasyOCR output on each source crop as a pseudo-reference, mean character error rate fell from 0.513 to 0.477, while exact-string agreement was unchanged (39.39\% vs. 39.34\%). Thus OCR-biased sampling modestly improves text-region DISTS and character-level recognizability, but the gain is small, does not improve exact recognition, and slightly hurts the overall challenge objective; this supports excluding it from the final candidate pool.

\paragraph{Conditioning refresh.} Because the autoregressive entropy context is derived from the latent being optimized, freezing it for the entire trajectory causes the context used to score candidate updates to grow stale as the latent moves. Periodically re-deriving the frozen context from the current latent every 75 iterations (\emph{refreeze}) was a strong development lever. An initial matched 300-step, 8-image screen was sign-mixed (6/8 improved, one $-1.02$ outlier, mean $-0.003$/image), whereas the full 100-image refreshed candidate improved its uncapped mean score by 0.145 at similar rate. The cumulative legal-rate submission changed from v17 to v20 by +0.136 $S$, but this historical transition also changed the candidate pool/selection procedure and is therefore not a clean causal ablation. Extending trajectory depth from 300 to 500 steps under the same refreeze schedule (v20$\to$v21) added a dev-set-estimated +0.17 $S$. These components were developed and frozen before the decoder submission. TTO is strictly an encoder-side operation: the submitted decoder performs no optimization, gradient computation, or online adaptation of any kind.

\subsection{Global MILP rate allocation}
\label{sec:knapsack}
Given candidate streams for image $i$ from checkpoint/refinement combination $j$, with challenge score $s_{ij}$ (computed against the source image, available to the encoder but not transmitted) and \emph{measured} coded length $r_{ij}$ bits, final per-image selection solves the multiple-choice knapsack
\begin{align}
 \max_{z_{ij}}\;&\sum_{i,j}s_{ij}z_{ij},\\
 \text{s.t. }&\sum_jz_{ij}=1\ \ \forall i,\quad
 \sum_{i,j}r_{ij}z_{ij}\leq0.025\sum_iH_iW_i,
\end{align}
with binary $z_{ij}$, solved with SciPy's \texttt{scipy.optimize.milp} (HiGHS branch-and-bound): one binary variable per (image, candidate), one equality row per image ($\sum_jz_{ij}=1$), one global inequality row on total bits, default solver tolerance, and a 600\,s time limit (never approached; solve time is a few seconds for the $\sim\!1000$ variables in our 10-candidate$\times$100-image pool). Since \cref{eq:score} is linear in the four per-image metrics, the organizers' aggregate score equals $\frac{1}{|N|}\sum_{i,j}s_{ij}z_{ij}$ exactly, so maximizing this objective is equivalent to maximizing the organizers' aggregate objective over the candidate pool. Two implementation details mattered: the constraint must use \emph{true pixel-weighted} BPP ($\sum r_{ij}z_{ij}/\sum H_iW_i$), not a mean of per-image BPP values, and $r_{ij}$ must be the actual on-disk file size including the header byte. Both $s_{ij}$ and $r_{ij}$ are measured before the solver runs. To isolate allocation, we retrospectively reran the retained final 10-candidate$\times$100-image test pool at an identical 0.02499-BPP cap. A deterministic ratio-greedy baseline---initializing each image at its smallest stream, then repeatedly taking the feasible positive-score replacement with greatest score gain per added bit---obtained $S{=}111.53236$ at 0.024976 BPP. HiGHS with zero relative MIP gap obtained $S{=}111.53643$ at 0.024990 BPP, a gain of $+0.00407$ $S$. This is a pool-level retrospective analysis, not a newly submitted result. It also showed that the submitted default-tolerance solution was near-optimal rather than certified optimal; accordingly, we make no stronger claim. The historical $+0.0375$ log used an unreconstructable baseline and is not used.

This is a full-reference, dataset-level encoder operation: it requires every source/candidate pair and the total pixel budget before solving. Once selected, each stream is independently decodable without its source or the rest of the dataset. An online or independently rate-controlled deployment would instead need a prescribed per-image target or a learned no-reference controller; the score-maximizing global allocation reported here should not be interpreted as such a method.

\subsection{Decoder-side residual restoration}
\label{sec:restorer}
The final stage is a fixed, approximately 1.5M-parameter residual CNN: a 64-channel head, eight dense-reuse residual blocks (three $3\times3$ convolutions/block, growth 32, residual scale 0.2), and a zero-initialized RGB tail. It predicts an RGB residual, after which the image is clamped to $[0,1]$. It consumes no source pixels or side information and therefore adds zero additional BPP. Its role is conservative correction of recurring color, edge, and local-texture error rather than unconstrained detail generation; on a separate 6-image subset (not the submitted TTO) we additionally routed TTO's gradient through this frozen restorer and found no improvement (mean $-0.036$ $S$, mixed signs, within noise), so this variant was rejected and the submitted TTO trajectory (\cref{sec:tto}) always optimizes against the raw, pre-restoration decode. See \cref{sec:ablation} for the restorer's independent, rate-orthogonal ablation.

\paragraph{Decoder consistency.} The official metrics in Table~4 were provided by the organizers from the submitted reconstruction archive. For 96 of 100 images, the archived reconstruction exactly matches the output of the submitted standalone decoder. For four images, encoder-side candidate selection retained the pre-restoration reconstruction, whereas the submitted decoder applies the fixed restorer unconditionally because the bitstream does not signal this choice. Consequently, those four archived PNGs are not reproduced exactly by the standalone decoder. All submitted bitstreams remain independently decodable. A one-bit restoration flag would eliminate this inconsistency in a subsequent implementation.

\subsection{Implementation and reproducibility}
The submitted decoder bundle contains the trimmed FP16 SD-Turbo UNet/VAE, the AdcSR half-decoder, the four AEIC-ME checkpoints, the restoration weights, the practical entropy-coding extension, inference scripts, dependency notes, and a README, totalling approximately 4.76\,GB unpacked, under the 5\,GB decoder-package limit. Because the decoder must be able to instantiate any of the four checkpoints depending on which byte it reads, all four codec networks (each including its own copy of the diffusion synthesis path) are resident on the GPU simultaneously during a decode run over a mixed-checkpoint bitstream directory; we found this leaves materially less headroom than a single-checkpoint decode, and a naive full-resolution application of the restoration network on top of this baseline memory footprint can exceed a 22--24\,GB GPU's budget on the largest ($>2048\times2048$) test images. Because the restoration network is a purely local, normalization-free convolutional stack, we apply it with overlapping-tile inference (tile size 512\,px, context padding 48\,px, comfortably larger than its $\approx$26-pixel receptive-field radius) so that peak activation memory is bounded independently of image resolution while producing output numerically identical to a single full-image forward pass. Reproduction requires Python 3.10, PyTorch 2.1.x with a CUDA 12.x-compatible runtime, CompressAI 1.2.8~\cite{begaint2020compressai}, Diffusers, xFormers, and the compiled rANS extension included by the build procedure.

\paragraph{Runtime.} The submitted archive metadata contains an 8.5\,s/image value, but this field was hard-coded during packaging and its scope was not recorded; we therefore report it as metadata, not as a measured encoder or decoder benchmark. Of the 10 candidate tags/image (\cref{sec:knapsack}), only one involves TTO: 4 raw checkpoint compressions plus their 4 zero-additional-bit restorer variants, and one 500-step refreeze-75 trajectory plus its restored copy (rf500/rf500\_enh share that trajectory). Direct logs cover 99 trajectories: TTO averaged 225.0\,s/image (212.7--245.1\,s) on one A10G and consumed 6.19 GPU-hours; including the cheaper passes, final archive generation took approximately 6.5 A10G-hours. Images can be sharded independently.

\section{Experiments}
\subsection{Training data}
The official benchmark, AIGC-IC-1000, contains 1000 images generated by 10 text-to-image models (100 images each) with prompts drawn from GenEval, DPG-Bench, and CVTG-2K (280 images each) and LongText-Bench (160 images), split 800/100/100 into train/validation/test; validation and test images preserve their original mixed 1K/2K resolutions. We used the 800 challenge-provided training images and additionally generated approximately 15{,}700 images from public GenEval, DPG-Bench, CVTG-2K, and LongText-Bench prompt collections using SDXL, SD3.5-medium, PixArt-Sigma, and Sana. This external corpus was used for codec fine-tuning; the final matched restorer was trained on four-rate reconstructions of the official 800-image training split (3200 pairs). No validation or test image updated model parameters. Validation was used for development/model selection, while test sources were used only by the permitted encoder-side latent optimization and global full-reference allocation. \textbf{Extra training data was used}. The packaged test ZIP's readme incorrectly recorded the extra-data flag as 0; this paper and the submitted factsheet give the correct disclosure.

\paragraph{Provenance.} We generated the synthetic corpus ourselves from the named public prompt collections. It does not reuse the organizers' generated images or generation seeds. Validation/test pixels were never used to update model weights; their permitted full-reference use for development and encoding is stated above.

\subsection{Training protocol}
\label{sec:training-protocol}
Codec fine-tuning started from the public AEIC-ME lineage and used AdamW with the staged learning rates recorded in the released YAML files (initially $2\times10^{-5}$). The four specialists share an architecture but differ in loss weights, initialization, and corpus (\cref{tab:ckpts}). The restorer used Adam with cosine decay from $2\times10^{-4}$, random patches up to $384\times384$, batch size 8, and unit-norm gradient clipping. Its adopted loss weights were L1:LPIPS:DISTS $=1:1:1.5$; edge-aware and adversarial variants were evaluated during development but did not improve the validation objective and were not adopted. Checkpoint/candidate selection used validation reconstructions and measured stream lengths, considering all four challenge metrics, BPP, and standalone-decoder reproducibility.

\subsection{Evaluation protocol}
\label{sec:eval}
The organizers' score combines all four metrics as
\begin{equation}
S=\mathrm{PSNR}+10\,\mathrm{MS\mbox{-}SSIM}+40(1-\mathrm{LPIPS})+40(1-\mathrm{DISTS}),
\label{eq:score}
\end{equation}
equivalently $S = \mathrm{PSNR} + 10\,\mathrm{MS\mbox{-}SSIM} - 40\,\mathrm{LPIPS} - 40\,\mathrm{DISTS} + 80$, i.e. $S$ is exactly the sum of the four raw metrics as written in the organizers' rules. We verified our local evaluation reproduces the organizers' \emph{displayed} leaderboard number digit-for-digit across every submission for which both were available before the final test phase, and found that displayed number to equal $B \triangleq S - 80$ in every case (e.g.\ our own official entry: $S=111.527739$, displayed $B=31.527739$; \cref{tab:results}). \textbf{Scale convention for this paper:} development tables (\cref{tab:e1e3,tab:ablation}) report the unshifted $S$, matching our development log; official leaderboard results (\cref{tab:results,tab:leaderboard}) and any number explicitly marked ``board''/$B$ report $B=S-80$, the organizers' displayed scale. Every table and in-text number states or is adjacent to its scale; where both appear together (e.g.\ \cref{sec:discussion}), each value is individually labeled. The organizers determine the final test-phase ranking from two perspectives: (i) this objective score, and (ii) a subjective assessment of content consistency and perceptual quality from human annotations; the awards are correspondingly split into two non-overlapping objective-performance awards and three perceptual-performance awards. \Cref{tab:leaderboard} reports the objective-score ranking, which is the only leaderboard released to participants at the time of writing; our system was optimized solely against \cref{eq:score} and was not tuned against any human-preference signal.

\begin{table}[t]
\centering\footnotesize
\begin{tabular}{@{}lccc@{}}
\toprule
Checkpoint & Nominal rate & Score & BPP \\
\midrule
ft8 (zero-shot) & low & 106.19 & 0.017 \\
ft4 (zero-shot) & mid & 109.21 & 0.0239 \\
ft2 (zero-shot) & high & 111.28 & 0.031 \\
Early zero-shot mix (mean-BPP solver) & -- & 109.55 & 0.0250 \\
\bottomrule
\end{tabular}
\caption{Zero-shot AEIC-ME rate ladder on the challenge validation set, before any AIGC fine-tuning, TTO, or restoration ($S$ scale, i.e. leaderboard scale $+80$; see \cref{sec:eval}). A single checkpoint at the legal rate already trails a per-image mix of unrelated rate points, motivating the checkpoint-bank design used throughout the rest of the system. Under the same early (mean-BPP-constrained, later superseded) solver, adding three fine-tuned checkpoints to this pool raised the mix from 109.55 to 109.89 (+0.34); this same-methodology signal fine-tuning was worth adding, not comparable to \cref{tab:ablation}'s later rows.}
\label{tab:e1e3}
\end{table}

\subsection{Main results}
\Cref{tab:results} reports our official final-test-phase metrics as recalculated and announced by the organizers on August 4, 2026. \Cref{tab:leaderboard} reproduces the organizer-announced final ranking for all ten verified entries. We rank 5th; the top four entries all report substantially higher PSNR (28.6--29.8\,dB against our 27.0\,dB) at comparable or tighter BPP, while our DISTS (0.038970) is second-lowest of all ten entries; only prophet1 (0.038851, ranked 6th) is lower, and the lowest among every entry ranked above us.

\begin{table}[t]
\centering\small
\begin{tabular}{@{}lc@{}}
\toprule
Quantity & Final test entry \\
\midrule
Official score ($B{=}S{-}80$) & 31.527739 \\
PSNR & 27.022356 dB \\
MS-SSIM & 0.917594 \\
LPIPS & 0.077794 \\
DISTS & 0.038970 \\
Average BPP & 0.024946 \\
Archive metadata runtime & 8.5 s/image \\
Execution device & GPU \\
\bottomrule
\end{tabular}
\caption{Official organizer-recalculated final-test metrics for the submitted entry.}
\label{tab:results}
\end{table}

\begin{table}[t]
\centering\scriptsize
\resizebox{\textwidth}{!}{%
\begin{tabular}{@{}clcccccc@{}}
\toprule
Rank & Team & Score & BPP & PSNR & MS-SSIM & LPIPS & DISTS \\
\midrule
1 & Ashes & 34.004381 & 0.025000 & 29.820286 & 0.930888 & 0.076335 & 0.051785 \\
2 & rrrrty & 33.557593 & 0.024987 & 28.926965 & 0.934522 & 0.071049 & 0.046815 \\
3 & NeFIC++ & 33.271115 & 0.024153 & 28.601395 & 0.932167 & 0.071724 & 0.044575 \\
4 & Freedom & 32.849392 & 0.025000 & 28.779010 & 0.933339 & 0.077461 & 0.054115 \\
\rowcolor{blue!8}
5 & \textbf{ZeroR} & \textbf{31.527739} & \textbf{0.024946} & \textbf{27.022356} & \textbf{0.917594} & \textbf{0.077794} & \textbf{0.038970} \\
6 & prophet1 & 30.349026 & 0.024962 & 26.303701 & 0.908165 & 0.087057 & 0.038851 \\
7 & qrchen & 30.183789 & 0.024948 & 27.740318 & 0.917277 & 0.109328 & 0.058905 \\
8 & Team-Team & 29.318431 & 0.023415 & 26.729861 & 0.915676 & 0.104364 & 0.059841 \\
9 & rohit\_choudhary & 28.221412 & 0.024785 & 27.014252 & 0.904016 & 0.126391 & 0.069434 \\
10 & aigilic & 27.979867 & 0.021782 & 26.129605 & 0.902043 & 0.114655 & 0.064599 \\
\bottomrule
\end{tabular}}
\caption{Official final test-phase leaderboard, organizer-recalculated and announced August 4, 2026, at full precision. Our DISTS (0.038970) is second-lowest overall (only prophet1's 0.038851 is lower) and lowest among every entry ranked above us; our LPIPS is not a leadership position. PSNR trails the top four by 1.6--2.8\,dB (\cref{sec:discussion}).}
\label{tab:leaderboard}
\end{table}

\FloatBarrier
\subsection{Qualitative results}
\begin{figure}[h!]
\centering
\begin{tabular}{@{}cccc@{}}
\scriptsize Source (GT) & \scriptsize Fine-tuned decode & \scriptsize $+$ 500-step TTO & \scriptsize $+$ fixed restorer \\
\includegraphics[width=.235\textwidth]{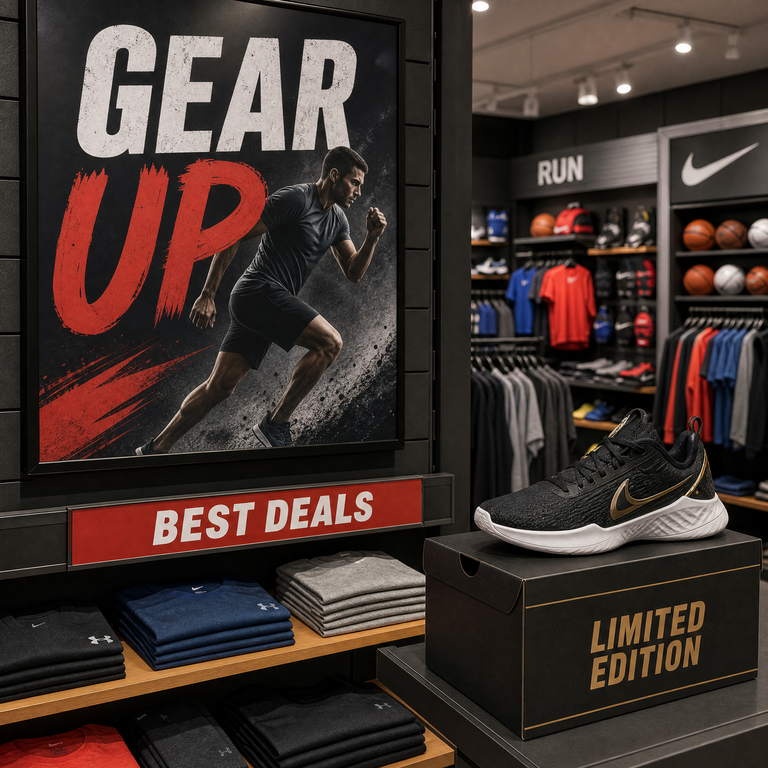} &
\includegraphics[width=.235\textwidth]{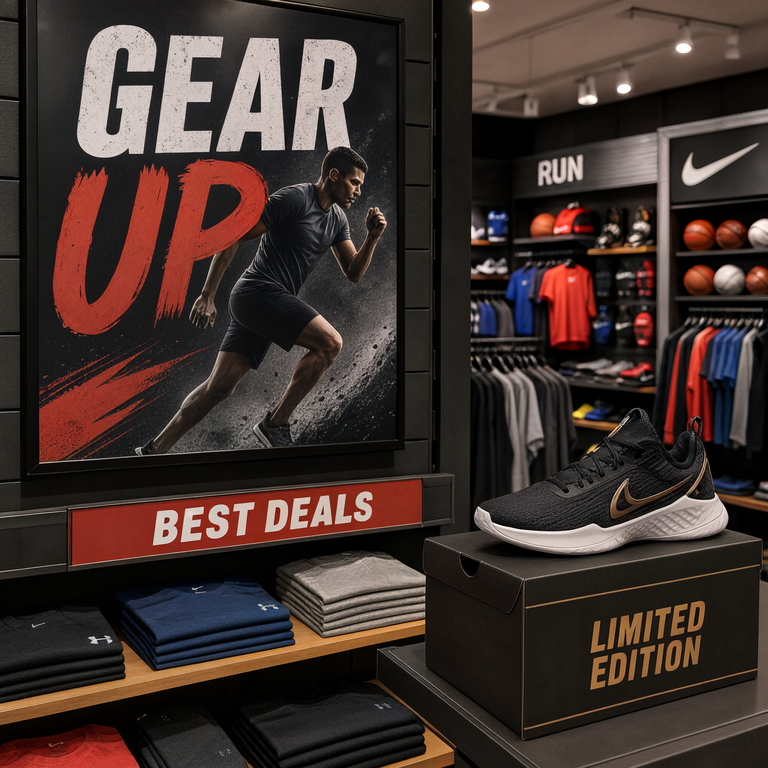} &
\includegraphics[width=.235\textwidth]{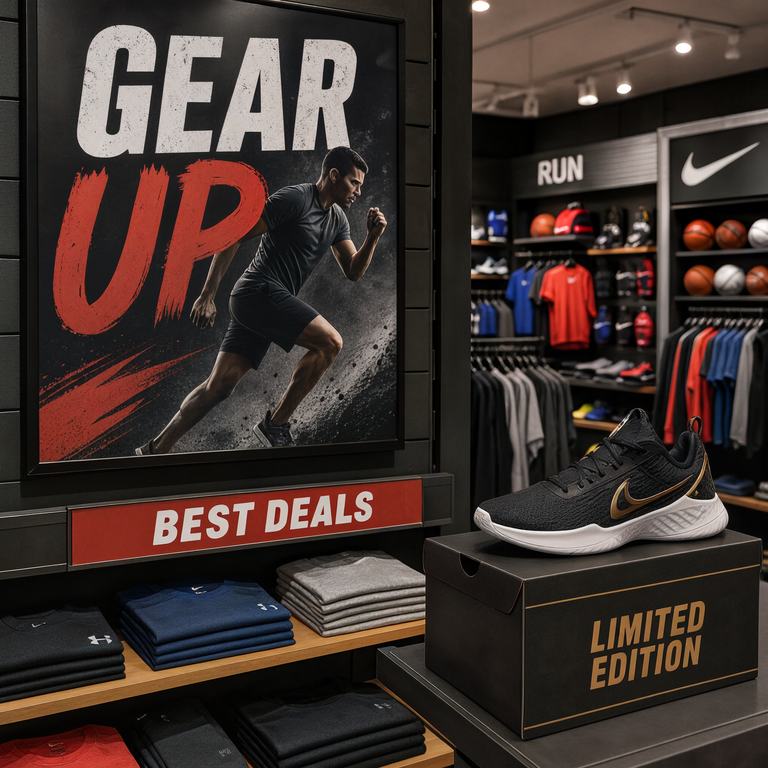} &
\includegraphics[width=.235\textwidth]{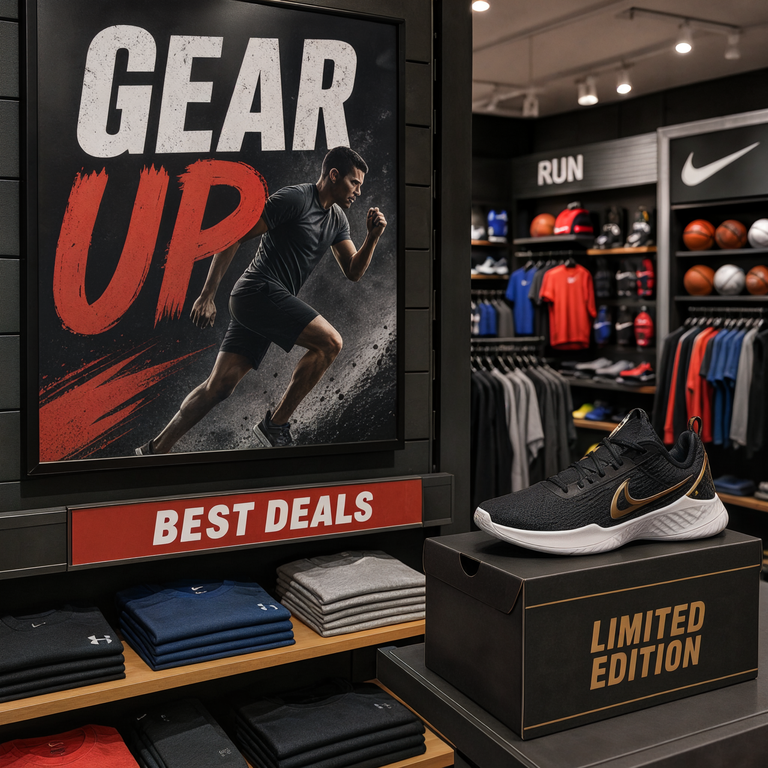} \\[2pt]
\multicolumn{4}{c}{\scriptsize Zoomed text crop (same coordinates in every column)} \\[-1pt]
\includegraphics[width=.235\textwidth]{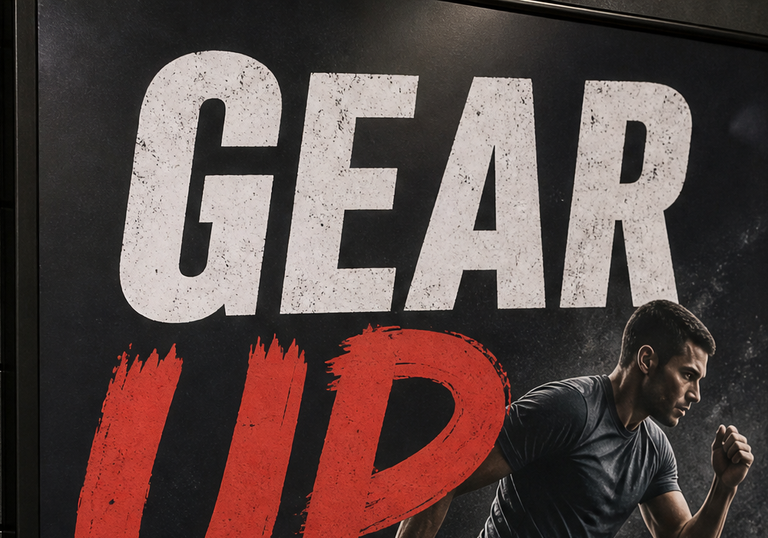} &
\includegraphics[width=.235\textwidth]{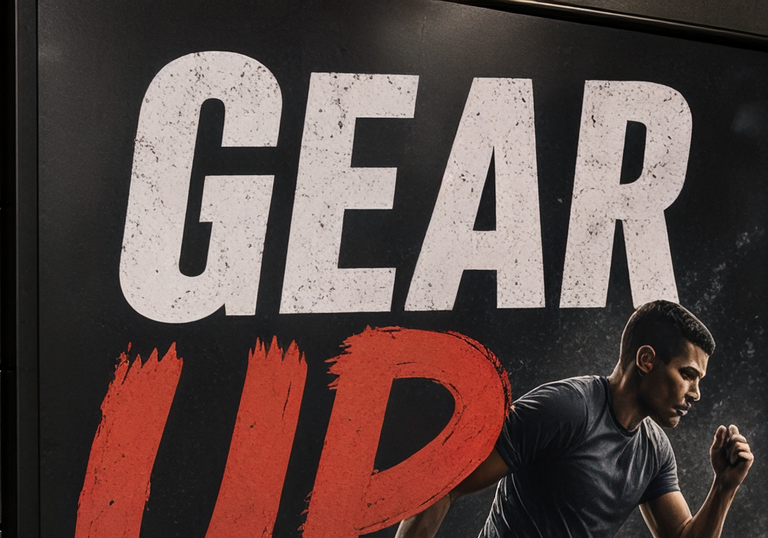} &
\includegraphics[width=.235\textwidth]{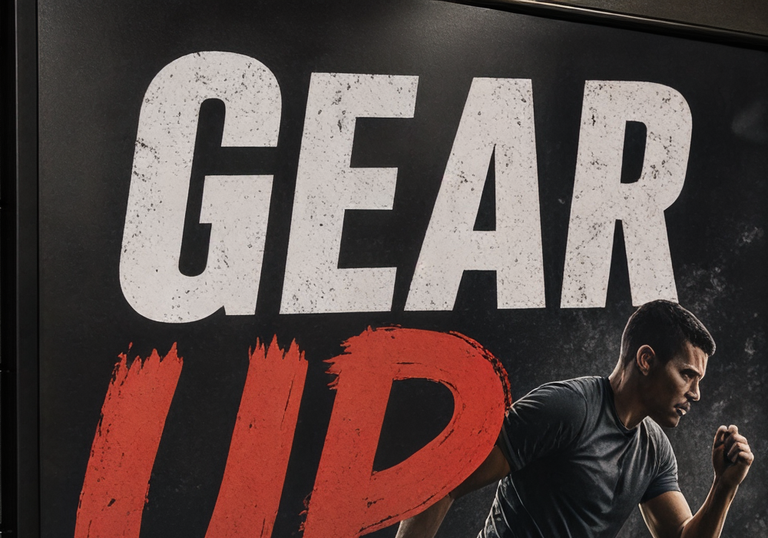} &
\includegraphics[width=.235\textwidth]{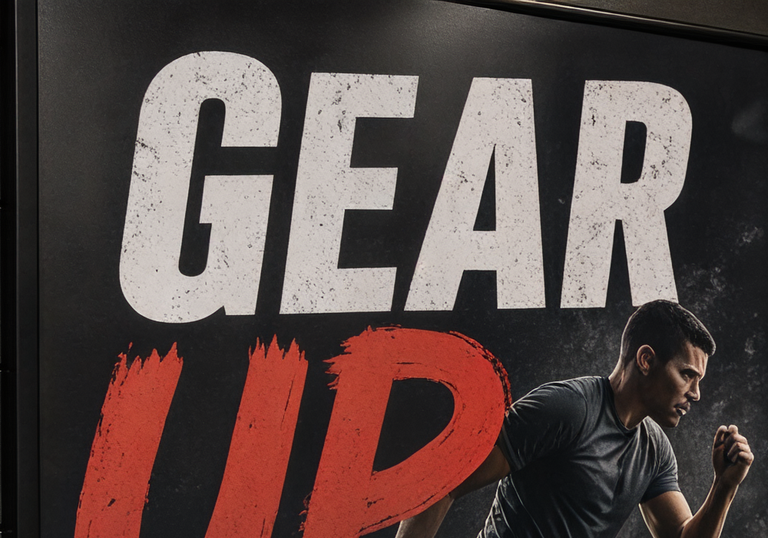}
\end{tabular}
\caption{Subjective stage comparison on hand-selected test image 000009; bottom row enlarges the same poster region. The fine-tuned decode is 0.01774 BPP; TTO produces the selected 0.01885-BPP stream; restoration adds no bits, and the last column is the submitted reconstruction. The comparison isolates visible stage effects, not aggregate performance; see \cref{sec:ablation} for full-set evidence.}
\label{fig:qual}
\end{figure}

\Cref{fig:qual} provides a stage-wise subjective comparison for test image 000009, a text-heavy retail-poster scene included among the 96 images whose archived reconstruction matches the standalone decoder. Relative to the frozen fine-tuned checkpoint decode, 500-step TTO sharpens the rendered typography and repeated garment/shoe structure while increasing the measured stream rate from 0.01774 to 0.01885 BPP; the fixed restorer then makes smaller edge and local-texture corrections at zero additional bits. The final column is the reconstruction from the actual submitted bitstream. We hand-selected this example because it exposes the system's intended text/structure behavior; quantitative claims rest on the full-set results in \cref{tab:leaderboard,tab:ablation}, not on this figure.

\subsection{Submission progression and development history}
\label{sec:ablation}
\Cref{tab:ablation} traces the system's development, drawn from 77 logged experiments spanning July 5--31, 2026. Progression rows are legal-rate mixed candidates, not raw operating points. The v3$\to$v17 jump jointly expands the fine-tuned checkpoint bank and adds 300-step TTO; v17 still used the historical remix, so this is not an isolated ablation. The v17$\to$v20 transition jointly adds refreshed TTO and the later MILP candidate-pool/selection procedure, so it is also not cleanly attributable. The matched rows separately isolate TTO (+0.3310) with the same MILP allocation at legal rate. Finally, v20$\to$v21 adds the 500-step refreshed candidate. Rows marked \emph{confirmed} were scored by the organizers' validation leaderboard; rows marked \emph{dev-set est.} were evaluated locally on validation. All scores are on the $S$ scale (\cref{sec:eval}).

\begin{table}[t]
\centering\scriptsize
\begin{tabular}{@{}lccl@{}}
\toprule
Candidate pool (cumulative) & $S$ & BPP & Status \\
\midrule
Early 11-way mixed pool (v3) & 109.94 & 0.0250 & confirmed \\
+ expanded fine-tuning, 300-step TTO (v17) & 111.53 & 0.0250 & confirmed \\
\quad + refresh, MILP pool (v20) & 111.66 & 0.0250 & confirmed \\
\quad\quad + 500-step TTO (v21) & 111.83 & 0.0250 & dev-set est. \\
\midrule
Matched pool without TTO & 111.18 & 0.0250 & dev-set ablation \\
Matched pool with TTO & 111.51 & 0.0250 & dev-set ablation \\
Official organizer-scored archive (test) & 111.53 & 0.0249 & confirmed (test) \\
\bottomrule
\end{tabular}
\caption{Legal-rate ($\leq0.025$ BPP) results. Progression and matched-ablation rows use validation; the last row is the official test set and is not directly comparable across datasets. In particular, v20 was not evaluated on test, so v20$\to$v21 transfer cannot be inferred from the official test score. The matched TTO comparison uses the same four checkpoints, restorer, and MILP allocation, differing only in the availability of TTO candidates.}
\label{tab:ablation}
\end{table}

\paragraph{Restoration network.} Because the fixed residual restorer (\cref{sec:restorer}) is applied post-decode with zero additional bits, its effect is rate-orthogonal and does not require a matched-rate comparison: applying it to reconstructions from every candidate checkpoint type gave a uniform +0.22 $S$ average gain (range +0.19 to +0.25 across candidate types, 100-image validation set), and extending its training from 8k to 18k steps on the same fixed L1+LPIPS+DISTS objective (no adversarial loss, no architecture change) raised this further to +0.27, confirming the gain came from training duration rather than loss-function or capacity changes. A 16-block (2$\times$-capacity) variant tested at matched convergence scored slightly \emph{worse} (111.25 vs.\ 111.27 $S$ on a fixed candidate set), closing out the architecture search.

\paragraph{Candidate-level screening (uncapped rate).} These, measured on raw reconstructions or small subsets, decided what to \emph{add} to the legal-rate pool in \cref{tab:ablation}, not standalone legal-rate claims: wider TTO crop context (\texttt{crop\_ly} 16$\to$24) was net-negative once re-mixed through the knapsack; MSE weight 1000$\to$300$\to$0 monotonically hurt score (29.88$\to$29.82$\to$29.79 $S{-}80$), so it stabilizes the search rather than only trading off perceptual terms; DISTS surrogate weight 40$\to$80$\to$120 didn't improve whole-image DISTS and damaged LPIPS/PSNR; multi-seed TTO was seed-insensitive; checkpoint weight-souping underperformed both parents; cross-family fusion with OSCAR~\cite{guo2025oscar}/OneDC~\cite{xue2025onedc} was non-competitive (\cref{sec:related}).

\section{Discussion: the PSNR gap}
\label{sec:discussion}
We investigated whether the observed PSNR gap to higher-fidelity competitors could be reduced by remixing existing candidates or re-weighting the TTO objective. Two experiments provide different strengths of evidence and are reported separately.

\paragraph{Full-test-set pool analysis ($n{=}100$).} Re-solving the existing ten-candidate test pool with PSNR weighted 1$\times$, 2$\times$, and 4$\times$ relative to the other metrics raised achievable PSNR at legal rate only to 27.08\,dB, a 0.06\,dB improvement, while combined score fell from $B{=}31.53$ to $31.48$. Thus, \emph{given the already-generated candidates}, selection re-weighting cannot materially close the PSNR gap. This analysis characterizes the fixed pool and is not evidence about unseen-data generalization.

\paragraph{Small-sample screen (preliminary, $n{=}4$).} We separately added an explicit, differentiable PSNR term ($10\log_{10}(1/\mathrm{MSE})$) directly into the TTO objective (\cref{eq:tto}) and re-ran the 500-step/refreeze-75 schedule on four representative test images: mean PSNR was 26.05\,dB against the existing rf500-enh candidate's 26.16\,dB on the same four images, with per-image deltas of $+0.057$, $-0.562$, $+0.121$, $-0.060$\,dB and simultaneous regression of the perceptual metrics. With $n{=}4$ and no dispersion estimate, this cannot by itself rule out that a different learning rate, weight schedule, or longer trajectory would recover the PSNR term's intended effect; we report it as a preliminary negative signal, not proof.

\paragraph{Interpretation.} Together with the full-pool result above and our observation that generic distortion-oriented codecs improved PSNR on selected images at the cost of substantially worse LPIPS/DISTS (why they were not used), we consider it \emph{likely} rather than proven that the residual PSNR gap reflects the one-step diffusion synthesis path itself, not a loss-weighting artifact correctable within our pool; closing it would plausibly need a higher-fidelity training objective or a smaller generative degrees-of-freedom budget. Under \cref{eq:score}, our confirmed strength is DISTS (second-best overall, best among the top five), not PSNR or LPIPS (\cref{tab:leaderboard}).

\section{Conclusion}
We presented our submission: four fine-tuned AEIC-ME checkpoints, encoder-side latent TTO with entropy-conditioning refresh, multiple-choice-knapsack MILP allocation over measured coded sizes, and a fixed zero-additional-bit restoration network, reaching rank 5 (score 31.527739, second-best DISTS) while trailing the top four on PSNR. Given the split award tracks (\cref{sec:eval}), the perceptual track fits better here. This gap likely reflects the diffusion backbone's fidelity/perception trade-off, not a tunable loss weight (\cref{sec:discussion}); fine-tuning or replacing the base codec's fidelity objective is the most promising path forward.

\paragraph{Compliance statement.} The described codec/checkpoints are those submitted for the Code Submission Phase; the official score corresponds to the submitted reconstruction archive; standalone-decoder consistency is discussed in Section~3.6. TTO does not modify decoder weights (pretrained components: \cref{sec:backbone}; extra data: Sec.~4.1).

\bibliographystyle{splncs04}
\bibliography{egbib}
\end{document}